\documentclass[11pt]{article}

\usepackage[final]{acl}

\usepackage{times}
\usepackage{latexsym}

\usepackage[T1]{fontenc}
\usepackage[utf8]{inputenc}

\usepackage{microtype}

\usepackage{inconsolata}

\usepackage{graphicx}
\usepackage{booktabs}
\usepackage{multirow}

\usepackage{booktabs}

\usepackage{tikz}
\usetikzlibrary{shapes, arrows.meta, positioning}
\usepackage{xcolor}

\title{Unreliable Narrator Framing as a Cost-effective Mechanism for Satire Explanation of Australian and American Satirical News}

\author{Kanay Gupta, David Y. Liu, Aditya Joshi \\
  University of New South Wales, Australia \\\texttt{kanay.gupta@student.unsw.edu.au, david.liu2@unsw.edu.au, aditya.joshi@unsw.edu.au}}

\begin{document}
\maketitle
\begin{abstract}
Satirical news is a unique form of misinformation: it is intended to be humorous by the original publication source but can be misinterpreted. Explanation of satirical news articles is challenging because recent and local context needs to be incorporated. This is a pilot study that explores explanation generation for satirical news articles from Australia and compares them alongside those from USA. We show that framing explanations with unreliable narrator theory on open-weight models can be a feasible, cost-effective alternative to frontier models. We find that models identify the satirical tropes more accurately on the American source than the Australian one. This work results in a novel dataset and an evaluation benchmark for satire explanation. 
\end{abstract}

\section{Introduction}
% Satire. Satire important in modern communication. Political and social commentary.

% Cite Yes But .... and in one sentence, its limitations. Memes. Satirical news articles.

% Culture/Demographic context. 
% One or two sentences about unreliable narrator.

% This is the first work that explores satire explanation for Australian news articles.
% 1-2 sentences description:
% 1 sentence on data collection and annotation
% We compared X models on using the unreliable narrator framework alongside simple baseline prompting.

% 1 sentence about broad observations.

Satirical news is a popular form of social media content which represents a unique form of misinformation. Unlike fake news, the intent of factually inaccurate satirical news is not deceptive. Satirical news functions as a means of delivering political and social commentary through humour and critique, but it is not always understood \citep{poulsen2023comparing}. Explaining satirical news and its real world commentary requires explaining what makes it satirical and the real-world phenomena it references. Previous work has shown that vision-language models struggle both to detect satire in images and to explain the source of its satirical effect \citep{nandy-etal-2024-yesbut}.

Betoota Advocate is an Australian news website that publishes satirical content, which typically references recent and geography-specific context which might be absent from a language model's training data. We collect 1000 articles from Betoota Advocate and The Onion, its American counterpart, and generate explanations of satirical news using four open-weight models and two frontier models.

% Consider the satirical headline ``Taylor Swift Makes Her Way Down The Aisle On Private Jet''\footnote{https://www.betootaadvocate.com/taylor-swift-makes-her-way-down-the-aisle-on-private-jet/}, understanding it requires knowing both that the singer married and that her private jet usage is widely criticised. Past work in satire detection and explanation attempts to address this peculiar form of news articles. 

% \citet{nandy-etal-2024-yesbut} reframed the task from satire detection to comprehension with YesBut, showing that vision language models can identify that an image is satirical while failing to explain why. Their dataset consists of cartoons, so the world knowledge required is largely visual and generic rather than tied to a geographical context.

% The two settings share the internet lookup scaffold, the output format and judge for the purposes of comparison. 

\begin{figure}[h]
\centering
\resizebox{0.8\columnwidth}{!}{%
\begin{tikzpicture}[
    font=\sffamily\scriptsize,
    >=Stealth,
    basecard/.style={
        rectangle,
        rounded corners=5pt,
        inner sep=5pt,
        line width=0.8pt,
        align=left
    },
    sourcecard/.style={
        basecard,
        draw=gray!50,
        fill=gray!5,
        text width=3.4cm
    },
    card1/.style={
        basecard,
        draw=blue!65!black,
        fill=blue!5,
        text width=3.6cm
    },
    card2/.style={
        basecard,
        draw=teal!65!black,
        fill=teal!5,
        text width=3.6cm
    },
    card3/.style={
        basecard,
        draw=violet!65!black,
        fill=violet!5,
        text width=3.6cm
    },
    connarrow/.style={
        ->,
        line width=1.2pt,
        draw=gray!60,
        shorten >=2pt
    }
]

% Middle Card (Reference anchor for vertical layout)
\node[card2] (c2) {
    \textbf{\textcolor{teal!75!black}{(2) Unreliable Narrator}}\\[2pt]
    Adopts the \textbf{naïf narrator trope}; interprets the cancellation as an earnest, unfortunate failure of commercial funding.
};

% Top Card
\node[card1, above=0.22cm of c2] (c1) {
    \textbf{\textcolor{blue!75!black}{(1) Satire Target}}\\[2pt]
    Sydney Mardi Gras cancels its major after-party event despite visible, widespread corporate involvement.
};

% Bottom Card
\node[card3, below=0.22cm of c2] (c3) {
    \textbf{\textcolor{violet!75!black}{(3) Implied Author}}\\[2pt]
    Exposes the \textbf{satirical gap}; satirizes the hypocrisy of performative ``rainbow capitalism'' that reaps brand profit without real support.
};

% Left Source Card
\node[sourcecard, left=0.65cm of c2] (source) {
    \textbf{Article Text} (\textit{The Betoota Advocate})\\[3pt]
    ``Sydney's Gay and Lesbian Mardi Gras is this week dealing with the bombshell revelation \textcolor{blue!75!black}{\textbf{their giant after party event won't be happening this year}}.\\[3pt]
    \textcolor{teal!75!black}{\textbf{Unfortunately}}, despite the likes of banks, tech companies and food giants \textcolor{violet!75!black}{\textbf{happily commandeering the pride flag}}.''
};

% Connecting Arrows
\draw[connarrow] ([yshift=0.55cm]source.east) -- (c1.west);
\draw[connarrow] (source.east) -- (c2.west);
\draw[connarrow] ([yshift=-0.55cm]source.east) -- (c3.west);

\end{tikzpicture}%
}
\caption{Unreliable narrator framing of an Australian satirical news article. The satire emerges from the unreliable narrator's surface account diverging from the implied author's critique of performative allyship.}
\label{fig:narrator_framing}
\end{figure}
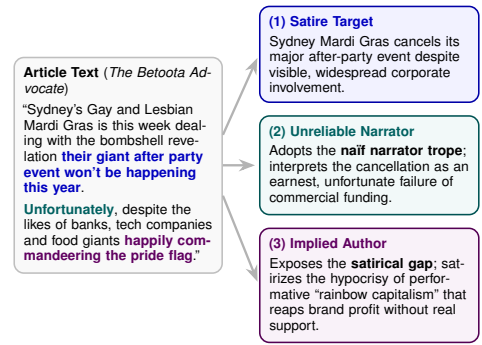

We apply \citet{booth1961rhetoric}'s unreliable narrator framing (Figure \ref{fig:narrator_framing}) through prompting used alongside internet search tool calling. We observe that the unreliable narrator prompt improves explanation quality in 16 of 17 comparisons with a baseline prompt, with the largest gains on the weakest models and no measurable advantage in the frontier models, suggesting that structured framing compensates for limited capability. 

Ours is the first work to evaluate satire explanation on Australian satirical news articles, and to use the literary concept of the unreliable narrator \citep{booth1961rhetoric} as a prompting framework for satirical news explanation generation (Figure \ref{fig:narrator_framing}). We make three observations. \textbf{(1)} Every model identifies the narrator trope more accurately on The Onion than on Betoota, a consistent difference between the two outlets. \textbf{(2)} The trope a model assigns reflects the model rather than the article, with each model applying its own preferred trope to both sources regardless of the text. \textbf{(3)} Performance degrades as more implicit interpretation is required, from describing the surface account, to inferring the implied author's belief, to explaining the gap between them.

\section{Related Work}
% 3/4th of half column.
% 5-6 papers on satirical explanation/classification.

% Explanation generation.
% LAst sentence: (a) Australian context, (b) Unreliable narrator framing

Work on satire has moved from binary classification toward explanation generation. Across that shift, the recurring difficulty is not recognising that a text is satirical but providing the world knowledge that explains why, and this becomes harder with cultural specificity. \citet{fakenewsvssatire} introduce a fake news versus satire dataset organised by the theme, but text-only classifiers trained on such data learnt the stylistic shortcuts rather than the content of the joke. \citet{li-etal-2020-multi-modal} showed that combining the headline and image features with ViLBERT outperforms unimodal baselines, though the model still cannot articulate what makes an article satirical. \citet{nandy-etal-2024-yesbut} reframed the problem as comprehension with YesBut, reporting that vision-language models score below 60\% accuracy on explanation generation for the satirical memes. Figurative language has also followed the same shift from labelling to explanation, as seen in FLUTE \cite{chakrabarty-etal-2022-flute} and sarcasm explanation in multi-party dialogue \cite{kumar-etal-2022-become}. \citet{abdalla-etal-2024-comparison} finds that chain-of-thought prompting recovers much of the performance lost on Arabic satirical news. No prior work evaluates satire explanation on Australian satirical news articles, and no prior work uses unreliable narrator \cite{booth1961rhetoric} as a prompting framework for explanation generation in the news domain.

\section{Methodology}

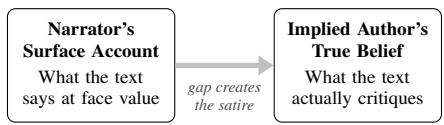
\begin{figure}[h]
\centering
\begin{tikzpicture}[
    box/.style={
        draw,
        rounded corners=3pt,
        text width=2cm,
        minimum height=1.5cm,
        align=center,
        inner sep=3pt,
        font=\scriptsize
    },
    arr/.style={
        -{Triangle[length=2mm, width=2.5mm]},
        line width=2pt,
        draw=gray!50
    }
]

% Left box
\node[box] (left) {
    \textbf{Narrator's Surface Account}\\[0.6ex]
    What the text says at face value
};

% Right box
\node[box, right=1.3cm of left] (right) {
    \textbf{Implied Author's True Belief}\\[0.6ex]
    What the text actually critiques
};

% Arrow with text
\draw[arr] (left) -- (right) 
    node[midway, below=2pt, align=center, font=\tiny\itshape, text=black!70] {
        gap creates\\the satire
    };

\end{tikzpicture}
\caption{Satirical gap model based on \citet{booth1961rhetoric}.}
\label{fig:satirical_gap}
\end{figure}

Given a satirical news article, the model is tasked with generating: (1) description of the \textbf{satire target} which could be a real-world event, person or phenomenon, and (2) an explanation of what is being satirised and how (Figure~\ref{fig:narrator_framing}). 

We hypothesise that satirical news uses the separation between the narrator of the text from its implied author, which is the set of values the article is actually trying to imply, in order to create the satire in the article (Figure~\ref{fig:satirical_gap}). We frame this implicit narrator in satire as an \textbf{unreliable narrator} \citep{booth1961rhetoric} which follows tropes such as the Naif who is naively blind to the obvious wrongness of what they describe and the Clown who reframes conflict or failure as normal or even positive \citep{brei-etal-2025-classifying} (Appendix \ref{appendix_tropes}). The model is then asked to state (1) which trope applies and why, (2) what the narrator presents on the surface, (3) what the implied author actually means, and (4) how the gap between the narrator and implied author creates the satirical effect. The detailed prompts are in Appendix \ref{appendix_prompts}.

Australian satire depends on knowledge such as local political figures, recent events or slang, which a model may not hold. To address this issue, the model is given access to perform web searches before answering (Figure~\ref{fig:internet}). Details about the implementation are given in the Appendix \ref{appendix_internet_search}. 

\begin{figure}[h]
\centering
\resizebox{\columnwidth}{!}{%
\begin{tikzpicture}[
    font=\sffamily\scriptsize,
    >=Stealth,
    node distance=6mm,
    io/.style={
        rectangle, rounded corners=4pt, draw=gray!55, fill=gray!5,
        line width=0.8pt, align=center, inner sep=5pt, text width=2.0cm
    },
    proc/.style={
        rectangle, rounded corners=4pt, draw=teal!65!black, fill=teal!8,
        line width=0.9pt, align=center, inner sep=5pt, text width=2.1cm
    },
    tool/.style={
        rectangle, rounded corners=4pt, draw=blue!55!black, fill=blue!6,
        line width=0.8pt, align=center, inner sep=5pt, text width=2.1cm
    },
    flow/.style={->, line width=1pt, draw=gray!65},
    loop/.style={->, line width=0.9pt, draw=blue!55!black}
]
 
\node[io] (in) {\textbf{Article}\\[1pt]headline\\+ body text};
\node[proc, right=9mm of in] (llm) {\textbf{Generator}\\[1pt]baseline or\\unreliable narrator};
\node[io, right=9mm of llm] (out)
    {\textbf{Output}\\[1pt]\texttt{REAL\_EVENT}\\\texttt{FINAL}};
\node[tool, below=9mm of llm] (web) {\textbf{DuckDuckGo}\\[1pt]top 3 results};
 
\draw[flow] (in)  -- (llm);
\draw[flow] (llm) -- (out);
 
\draw[loop] ([xshift=-6mm]llm.south) -- ([xshift=-6mm]web.north)
    node[midway, left=1pt, font=\tiny\itshape, text=blue!55!black]
    {\texttt{SEARCH:}};
\draw[loop] ([xshift=6mm]web.north) -- ([xshift=6mm]llm.south)
    node[midway, right=1pt, font=\tiny\itshape, text=blue!55!black]
    {title + snippet};
 
\node[below=1.5mm of web, font=\tiny\itshape, text=black!60]
    {at most 4 retrieval turns};
 
\end{tikzpicture}%
}
\caption{Both prompting conditions share the same retrieval scaffold. The model
may emit a \texttt{SEARCH:} directive and receive the top three DuckDuckGo
results before producing its \texttt{REAL\_EVENT} and \texttt{FINAL} blocks.}
\label{fig:internet}
\end{figure}
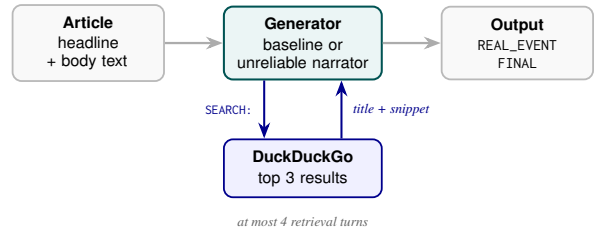

Additionally, we test a pre-processing step described in Appendix \ref{appendix_preprocessing} but find that it degrades explanation of the satirical target in the majority of cases so we do not use it in the main comparison.

\section{Experiment Setup}
% Names of models. Parameters
% Baseline: Prompting.
% Sentence classifiers.

We use 500 articles from The Betoota Advocate (\url{www.betootaadvocate.com}) and 500 from The Onion (\url{www.theonion.com}). Each record in the dataset includes a headline and body text. 

We evaluate four open-weight instruction-tuned models: Qwen2-VL-7B-Instruct, Qwen2.5-7B-Instruct, Llama-3.1-8B-Instruct and Mistral-7B-Instruct-v0.3. The Qwen2.5-7B model isolates the effect of removing vision within a single family while Llama and Mistral test whether the structured prompting has an effect across different families. As a comparison we additionally run GPT-5.6 Luna and GPT-5.6 Sol over 100 Betoota articles. 

We score the generations with Gemini 3 Flash using a rubric \citep{pombal2025mprometheus} in which the judge first writes feedback for the task and then assigns a score. All conditions are scored on satire target (1-5, accuracy of the target, \textit{i.e.}, the identified real-world event, person or phenomenon being satirised) and explanation (1-5, accuracy and completeness of the satire mechanism) (Appendix \ref{appendix_llm_judge}). UN outputs are additionally scored on four dimensions: trope accuracy (0/1, whether the judge finds the assigned trope well justified by the article), surface account, implied author and mechanism (each 1-5). We compare all judge scores with human ratings in Section~\ref{sec:human}.

\section{Results}
% 1 Table.
% Table X shows the performance of.......
% 1-2 sentences: "Read" numbers.
% 1-2 sentences: Trend.
% 1 Diagram. 

The unreliable narrator (UN) prompt improves explanation quality for every generator on both datasets (Table \ref{tab:main}). Across all the comparisons between the two datasets, four models and both sentence classification settings (Appendix \ref{appendix_preprocessing}), 16 out of 17 have a positive change when using the UN framework. Mistral on Betoota with sentence classification is the single negative outcome. Gains are largest where the baseline performance is the weakest. 

Satire target identification also shows improvement as 13 of the 17 outcomes favour the UN prompt. Roughly half of the outputs score a 1 on this dimension. Upon closer analysis of individual outputs, we found that the model frequently restates the article's invented characters and events as though they were literal fact rather than identifying the real world phenomenon the article references (see Appendix \ref{appendix_qualitative}). Generations without a well-formed \texttt{FINAL} block are excluded. These are almost entirely from Qwen2-VL-7B, and keeping them or scoring them as 1 does not change the direction of any explanation comparison.

\begin{table}[h]
\centering
\footnotesize
\setlength{\tabcolsep}{3pt}
\begin{tabular}{llcccc}
\toprule
 & & \multicolumn{2}{c}{\textsc{Betoota} (AU)} & \multicolumn{2}{c}{\textsc{The Onion} (US)} \\
\cmidrule(lr){3-4}\cmidrule(lr){5-6}
Model & Prompt & \textit{target} & \textit{expl.} & \textit{target} & \textit{expl.} \\
\midrule
\multirow{2}{*}{Qwen2-VL-7B}
 & \textsc{base} & \textbf{2.00} & 2.00 & \textbf{2.01} & 1.86 \\
 & \textsc{un}   & 1.93 & \textbf{2.46} & 1.96 & \textbf{2.38} \\
\addlinespace[2pt]
\multirow{2}{*}{Mistral-7B}
 & \textsc{base} & 1.76 & 2.75 & 2.09 & 2.45 \\
 & \textsc{un}   & \textbf{1.97} & \textbf{3.02} & \textbf{2.14} & \textbf{3.02} \\
\addlinespace[2pt]
\multirow{2}{*}{Qwen2.5-7B}
 & \textsc{base} & 2.03 & 2.31 & \textbf{2.28} & 2.19 \\
 & \textsc{un}   & \textbf{2.14} & \textbf{3.06} & \textbf{2.28} & \textbf{3.09} \\
\addlinespace[2pt]
\multirow{2}{*}{Llama-3.1-8B}
 & \textsc{base} & 1.68 & 2.83 & 2.01 & 2.91 \\
 & \textsc{un}   & \textbf{1.97} & \textbf{3.57} & \textbf{2.33} & \textbf{3.80} \\
\midrule
\multirow{2}{*}{GPT-5.6 Luna$^\ast$}
 & \textsc{base} & 4.08 & 4.91 & -- & -- \\
 & \textsc{un}   & \textbf{4.14} & \textbf{5.00} & -- & -- \\
\addlinespace[2pt]
\multirow{2}{*}{GPT-5.6 Sol$^\ast$}
 & \textsc{base} & 4.48 & 4.98 & -- & -- \\
 & \textsc{un}   & \textbf{4.54} & \textbf{5.00} & -- & -- \\
\bottomrule
\end{tabular}
\caption{Mean judge scores (1--5) under the baseline (\textsc{base}) and unreliable-narrator (\textsc{un}) prompts. Open-weight models were run with $n \leq 500$, generations that did not produce a well-formed "FINAL" block are excluded. $^\ast$Frontier models were run on Betoota only with $n=100$.}
\label{tab:main}
\end{table}

Table \ref{tab:fourdim} reports the four unreliable narrator dimensions which apply only to the UN condition. Trope identification is the weakest component which ranges from 0.25 to 0.74 for the open-weight models while the surface account is stronger, ranging from 3.37 to 4.53. We see performance degrade where additional inference is required beyond the literal content from the article, such as from surface account to implied author, then to satirical mechanism.

\begin{table}[h]
\centering
\footnotesize
\setlength{\tabcolsep}{4pt}
\begin{tabular}{lcccc}
\toprule
Model & \textit{trope} & \textit{surface} & \textit{author} & \textit{mech.} \\
\midrule
\multicolumn{5}{l}{\textit{Betoota Advocate} (AU)} \\
Qwen2-VL-7B         & 0.29 & 3.37 & 2.54 & 2.27 \\
Mistral-7B          & 0.25 & 3.94 & 3.55 & 2.84 \\
Qwen2.5-7B          & 0.35 & 3.80 & 3.16 & 2.75 \\
Llama-3.1-8B        & \textbf{0.46} & \textbf{3.88} & \textbf{3.87} & \textbf{3.25} \\
\addlinespace[2pt]
GPT-5.6 Luna        & 0.95 & 4.96 & 5.00 & 5.00 \\
GPT-5.6 Sol         & 0.98 & 5.00 & 4.94 & 5.00 \\
\midrule
\multicolumn{5}{l}{\textit{The Onion} (US)} \\
Qwen2-VL-7B         & 0.51 & 3.68 & 2.49 & 2.38 \\
Mistral-7B          & 0.33 & 4.07 & 3.45 & 2.74 \\
Qwen2.5-7B          & 0.54 & 4.36 & 3.06 & 3.10 \\
Llama-3.1-8B        & \textbf{0.74} & \textbf{4.53} & \textbf{3.83} & \textbf{3.77} \\
\bottomrule
\end{tabular}
\caption{Mean scores on the four unreliable-narrator (\textsc{un}) dimensions: \textit{trope} accuracy (0--1), \textit{surface} accuracy, implied \textit{author} belief, and
satirical \textit{mech}anism (each 1--5). Bold marks the best open-weight model per column.
Frontier models were run on Betoota only, $n{=}100$.}
\label{tab:fourdim}
\end{table}

Trope identification is where the sources diverge the most. Every model identifies the narrator trope more accurately on The Onion than on Betoota, by 0.08 to 0.28 (Table \ref{tab:fourdim}). The distributions models assign are consistent within a model but inconsistent across models. Llama-3.1-8B and Qwen2.5-7B label the majority of articles Naif in both sources, while Mistral-7B labels the majority Madman in both (Appendix~\ref{appendix_tropes}). Since all three receive identical inputs, this suggests a model-specific preference rather than a feature of the text, which is consistent with the low trope accuracies in Table~\ref{tab:fourdim}. The higher trope accuracy on The Onion may reflect greater model familiarity with American satire, although publication style and topic mix may also play a role. The gap between the frontier and open-weight models suggests that the bottleneck lies in recovering the real-world reference rather than in the framing or evaluation of the task.

\subsection{Human Evaluation}
\label{sec:human}

To validate the LLM as a judge, we sampled 50 Llama-3.1-8B responses on Betoota, stratified across both prompts and all five judge score bands. Three annotators scored them against the rubric in Table~\ref{tab:satire_rubric} with the same information available to the judge. Table~\ref{tab:human} reports agreement among the annotators and between their aggregated scores and the judge. Agreement is good for satire target (ICC$(2,k) = 0.78$) and moderate for mechanism and implied author (0.52 and 0.49), but poor for explanation (0.36), trope (0.22) and surface account ($-0.11$). The aggregated human score correlates significantly with the judge on five of the six dimensions ($\rho = 0.44$ to $0.76$), with the exception being surface account. The judge is stricter than the annotators on every dimension which is consistent between both prompts. We therefore rest our main claims on satire target and explanation and treat surface account and trope as indicative. 

\begin{table}[h]
\centering\small\setlength{\tabcolsep}{4pt}
\begin{tabular}{lccccc}
\toprule
Dimension & $n$ & human & judge & ICC$(2,k)$ & $\rho$ \\
\midrule
\textit{satire target} & 50 & 2.70 & 1.98 & 0.78 & 0.68$^{*}$ \\
\textit{explanation} & 48 & 3.24 & 3.00 & 0.36 & 0.68$^{*}$ \\
\textit{trope} & 24 & 0.70 & 0.44 & 0.22 & 0.76$^{*}$ \\
\textit{surface} & 24 & 4.01 & 3.64 & -0.11 & 0.30 \\
\textit{implied author} & 23 & 3.49 & 3.36 & 0.49 & 0.62$^{*}$ \\
\textit{mechanism} & 23 & 3.40 & 2.84 & 0.52 & 0.44$^{*}$ \\
\bottomrule
\end{tabular}
\caption{Human evaluation. \emph{human} is the mean of three annotators; ICC$(2,k)$ is the two-way random effects intraclass correlation for that mean; $\rho$ is Spearman correlation between the aggregated human score and the judge, with $^{*}$ marking $p<0.05$.}
\label{tab:human}
\end{table}

\section{Discussion}
\label{sec:discussion}
% Give a few examples that went right and that went wrong explain.
% Areas of improvement???

% suggesting that recognising the narrative trope depends on cultural register. 

% For the Betoota headline in Table~\ref{tab:score_difference_comparison}, the baseline prompt produced an explanation which scored 2 out of 5. It correctly identified the target as the hypocrisy from corporations and then stopped, whereas the unreliable narrator prompt scored a 5 as it named the Naif trope, described the surface account as a failure of the corporate sponsorships, stated the implied author's critique of the performative support from the corporations and then explained the gap between them. The gain came from the structure the second response was required to fill, not from new knowledge. Similar results were obtained for an Onion article (See Appendix). The gap between frontier and small open-weight models shows that the bottleneck is in the grounding rather than in the framing. For future work, a component that verifies whether the named entity exists before the model outputs it to the satire target field would fix incorrectly mistaking fictional beings as real.

\begin{table}[h]
\footnotesize
\centering
\begin{tabular}{@{} p{0.94\linewidth} @{}}
\toprule
\textbf{Input Headline:} ``Girlfriend Takes A Pause From Scrolling To Declare She’s Going To Read 20 Books This Year'' \\
\midrule
\textbf{Baseline Explanation (2 / 5).} The satire is about the superficial nature of
New Year’s resolutions and the tendency to prioritize social
media over personal growth. \\
\addlinespace[4pt]
\textbf{UN Explanation (5 / 5).} ...The surface account presents
Bianca Sutton as a woman who has decided to break away
from her TikTok addiction and read 20 books this year. However, the implied author, who is likely a satirist, actually
sees this as a continuation of her usual behaviour, where she
spends most of her time on social media. The gap between
the narrator’s presentation of a serious change in behaviour
and the implied author’s recognition of it as a superficial and
ultimately unchanged pattern creates the satirical effect. \\
\bottomrule
\end{tabular}
\caption{An example where a baseline prompt fails to produce a response that explains the satire.}
\label{tab:example}
\end{table}

The unreliable narrator prompt gives small models a structure to fill. Under the baseline prompt they tend to state that satire is present and name its target without explaining how the article produces it, whereas the UN prompt explains the satire by naming the trope, the surface account, the implied author's belief and the gap between them (Table~\ref{tab:example}). A control prompt with the same four steps but no narrator terminology gives similar or larger gains (Appendix~\ref{appendix_structured}), so this structure is a key ingredient of the framing. Since satirical articles are published continuously and at volume, a small open-weight model with a structured prompt offers a cost-effective way to produce explanations at scale.

Both frontier models more than double the satire target score under identical prompts, scaffolding and judge, showing that the rubric can award high scores and that the low ceiling reflects the capability of open-weight models. The UN framing gives no clear improvement at this scale given the lack of headroom, suggesting that structured framing compensates for part of the limited capability of small models on explanation, \textbf{which makes it a cost-effective alternative to frontier models}.

\section{Conclusion}
% Summary.
% Past tense.
% 6 sentences.
% Future work.

We evaluated whether formulating satirical news explanation through the framework of an unreliable narrator is beneficial. We compared a structured unreliable narrator prompt against a baseline explanation on 500 Australian and 500 American satirical articles. 

The unreliable narrator prompt improved explanation quality in 16 of 17 comparison combinations, where the gains were largest for the weakest models. Trope identification was consistently more accurate on the American source than the Australian one for every model tested, which may reflect greater familiarity with American satire. At the frontier scale, the advantage of the unreliable narrator framing disappeared, but the satire target scores more than doubled under identical conditions showing that the grounding was limited by model capability rather than the evaluation rubric. Future work should target the failures in grounding directly through verification during retrieval and should test whether framing transfers to other culturally specific satirical publications and to multimodal input.

\section*{Limitations \& Ethical Considerations}
\textbf{What drives the gain.} The baseline and UN prompts differ in task decomposition, instruction specificity and terminology, not only in theoretical framing. In follow-up experiments, a structured prompt with the same four steps but without narrator terminology gave larger explanation gains than UN in 7 of 8 cells (Appendix~\ref{appendix_structured}). The benefit therefore appears to come largely from decomposing the task. The narrator framing remains useful for its interpretable intermediate outputs (trope, surface account, implied author), which show where explanations fail.

\textbf{Evaluation.} All scores come from a single proprietary judge (Gemini 3 Flash), and we do not test whether the results hold under other judges. The judge sees the article and the answer but no verified reference material, so it cannot fully check real-world claims about recent or local events. Structured answers are longer and use the requested vocabulary, which the judge may reward independently of understanding. Human validation covers 50 responses from one model on one source. Agreement is weak for explanation and trope and negative for surface account. Trope accuracy has no gold label and is the judge's own assessment.

\textbf{Retrieval.} Both prompts share the search scaffold, but different prompts can trigger different searches, so the conditions may not see the same evidence. The frontier models mostly answered without searching. A no-retrieval ablation would separate the framing effect from retrieval.

\textbf{Publications and culture.} We use one publication per country, so differences between Betoota and The Onion cannot be separated into cultural effects and publication effects such as style, political leaning, topic mix or difficulty. Several publications per country (e.g., The Babylon Bee alongside The Onion) are needed for cultural claims.

\textbf{Ethics.} All news articles are publicly available. All annotators are authors of the work.

\bibliography{custom} 

\section*{Appendix}
\appendix

\section{Unreliable Narrator Tropes}
\label{appendix_tropes}
We take the unreliable narrator tropes described by \citet{brei-etal-2025-classifying} and \citet{riggan1978picaros}:

\begin{itemize}
\itemsep0em
\item \textbf{Naif}: naively blind to the obvious wrongness of what
they describe. The narrator reports a troubling situation in an
earnest, matter-of-fact register, and the satire comes from the
reader seeing what the narrator does not.
\item \textbf{Clown}: reframes conflict or failure as normal or even
positive. The narrator acknowledges that something has gone wrong
but treats it as unremarkable or as a cause for optimism.
\item \textbf{Picaro}: treats a cunning or self-serving act as
completely reasonable. The narrator presents self-interest, evasion
or exploitation as ordinary and justified conduct.
\item \textbf{Madman}: reports with excessive, disproportionate
emotional investment. The narrator's intensity is wildly out of
scale with the triviality of the subject.
\end{itemize}

\begin{table}[h]
\centering\small\setlength{\tabcolsep}{4pt}
\begin{tabular}{llcccc}
\toprule
Model & Source & Naif & Clown & Madman & Picaro \\
\midrule
\multirow{2}{*}{Llama-3.1-8B}
 & AU & 354 & 65 & 47 & 34 \\
 & US & 411 & 39 & 37 &  3 \\
\addlinespace[2pt]
\multirow{2}{*}{Qwen2.5-7B}
 & AU & 307 & 82 & 77 & 31 \\
 & US & 347 & 56 & 76 & 20 \\
\addlinespace[2pt]
\multirow{2}{*}{Mistral-7B}
 & AU &  35 & 68 & 295 & 98 \\
 & US &  44 & 92 & 343 & 21 \\
\bottomrule
\end{tabular}
\caption{Narrator tropes assigned under the unreliable narrator prompt, $n=500$ per source. Each model applies a consistent preference across both sources, but the preferences differ between models.}
\label{tab:tropes}
\end{table}
\section{Prompts}
\label{appendix_prompts}

We compare two methods, both of which share the same basic scaffold and output, described as follows: (A) \textbf{Baseline:} The prompt says that the model is an expert at analysing satire in news articles and asks for "a straightforward explanation of what the article is satirising." (B) \textbf{Unreliable Narrator (UN):} The prompt says the model is an expert in narratological analysis and supplies a background paragraph. The paragraph defines unreliable narrator and the four tropes. 

The two methods compared in this work are matched in that they must output a "REAL\_EVENT:" block followed by a "FINAL:" block. The specification for the "REAL\_EVENT:" is identical between the baseline and the unreliable narrator prompting and they only differ in the specification for the "FINAL:" block. The full prompts are as follows:

\paragraph{Baseline prompt.}
\begin{small}
\begin{verbatim}
You are an expert at analyzing satire in news
articles. You have access to a web search tool
to look up current events, politicians,
celebrities, or slang required to understand
the joke.
 
STRICT RULES:
1. If you need context, output EXACTLY:
   'SEARCH: [short keyword query]'. Use short
   keywords, NOT full sentences.
2. If you output SEARCH, STOP WRITING
   immediately. Do NOT add anything else. I will
   reply with the results.
3. Once you have enough context, output BOTH of
   the following - in this order, in the SAME
   turn, with no other text:
   REAL_EVENT: [A factual 1-3 sentence
   description of the real-world event, person,
   or phenomenon the article is referencing. Be
   specific - name the actual people, dates, or
   events involved.]
   FINAL: [A straightforward explanation of what
   the article is satirising.]
4. NEVER output SEARCH in the same turn as
   REAL_EVENT or FINAL.
\end{verbatim}
\end{small}
 
\paragraph{Unreliable narrator prompt.}
\begin{small}
\begin{verbatim}
You are an expert at analyzing satire using
narratological theory. You have access to a web
search tool to look up current events,
politicians, celebrities, or slang required to
understand the joke.
 
BACKGROUND: Satirical news sites write as an
'unreliable narrator' - a journalist who
presents absurd or exaggerated events with a
straight face. The narrator's surface account
diverges from the implied author's true intent,
and this gap IS the satire. The narrator
typically fits one of these tropes:
  - Naif: naively blind to the obvious wrongness
    of what they describe
  - Clown: reframes conflict or failure as normal
    or even positive
  - Picaro: treats a cunning or self-serving act
    as completely reasonable
  - Madman: reports with excessive,
    disproportionate emotional investment
 
STRICT RULES:
1. If you need context, output EXACTLY:
   'SEARCH: [short keyword query]'. Use short
   keywords, NOT full sentences.
2. If you output SEARCH, STOP WRITING
   immediately. Do NOT add anything else. I will
   reply with the results.
3. Once you have enough context, output BOTH of
   the following - in this order, in the SAME
   turn, with no other text:
   REAL_EVENT: [A factual 1-3 sentence
   description of the real-world event, person,
   or phenomenon the article is referencing. Be
   specific - name the actual people, dates, or
   events involved.]
   FINAL: [Analyse the satire through the
   unreliable narrator lens: (1) identify which
   narrator trope applies and why, (2) state what
   the narrator presents on the surface, (3)
   state what the implied author actually
   believes, and (4) explain how the gap between
   narrator and implied author creates the
   satirical effect.]
4. NEVER output SEARCH in the same turn as
   REAL_EVENT or FINAL.
\end{verbatim}
\end{small}
 
The user turn supplies the article headline and body text. The sentence
classification condition additionally prepends a note explaining the
\texttt{REAL} and \texttt{EXAGGERATED} labels applied to each body sentence.

\section{Internet Search}
\label{appendix_internet_search}

The model is given access to perform web searches. The model outputs "SEARCH:" along with a keyword, and the system returns the top three DuckDuckGo results with a title and a snippet, and then model continues. The model is able to do a total of 4 search loops before outputting the "REAL\_EVENT" and "FINAL" blocks. We refer to the "REAL\_EVENT" block as the satire target, as it is not always an event but the block name is unchanged from our earlier runs. 

\section{Preprocessing}
\label{appendix_preprocessing}
Qwen2-VL-7B-Instruct is prompted as the classifier to generate a JSON array of sentence and label pairs which was applied identically to both baseline and unreliable narrator prompting. The tagged body is substituted for the raw body in the generation prompt, along with a note explaining the labels in order to test whether the pre-processing improved the results.

The pre-processing raises the explanation scores under the baseline prompt in 5 of 8 comparisons (mean + 0.13 with the largest being Mistral-7B on Betoota at +0.38) but has essentially no effect under the unreliable narrator prompt (mean -0.02). Satire target scores, however, degrade consistently as they fall in 14 of 16 comparisons (mean -0.235) (Table \ref{tab:sc}). Labelling a sentence "EXAGGERATED" appears to discourage the model from using the information for the satire target even when a real event is referenced in the exaggerated sentence. Therefore, we do not use the sentence classifier pre-processing in the main comparison.

\section{Structured Control Prompt}
\label{appendix_structured}

To test whether the UN gain comes from the narrator framing or from the decomposition, we ran a structured control prompt (\textsc{struct}) on all four open-weight models and both sources (no sentence classification). It uses the baseline system prompt and search rules, and replaces only the \texttt{FINAL} specification with the UN four-step structure, with the narrator background, tropes and narrator/implied-author vocabulary removed:

\begin{small}
\begin{verbatim}
FINAL: [Analyse the satire step by step:
(1) identify the main satirical technique the
article uses and why, (2) state what the
article presents on the surface, (3) state
what the writer actually believes, and (4)
explain how the gap between what is presented
and what the writer believes creates the
satirical effect.]
\end{verbatim}
\end{small}

Table~\ref{tab:structured} compares it with the baseline and UN prompts. \textsc{struct} improves explanation over the baseline in all 8 cells and scores higher than UN in 7 of 8, the exception being Llama-3.1-8B on The Onion. On satire target the three prompts are close and the differences are mixed. \textsc{struct} answers are the longest of the three, so they carry the same length confound as UN. The result indicates that the narrator terminology is not what drives the explanation gain, but it does not show that \textsc{struct} reflects better understanding.

\begin{table}[h]
\centering
\footnotesize
\setlength{\tabcolsep}{2.5pt}
\begin{tabular}{llcccccc}
\toprule
 & & \multicolumn{3}{c}{\textit{target}} & \multicolumn{3}{c}{\textit{expl.}} \\
\cmidrule(lr){3-5}\cmidrule(lr){6-8}
 & Model & \textsc{base} & \textsc{un} & \textsc{str} & \textsc{base} & \textsc{un} & \textsc{str} \\
\midrule
\multirow{4}{*}{\rotatebox{90}{\scriptsize Betoota}}
 & Qwen2-VL & 2.00 & 1.93 & 2.07 & 2.00 & 2.46 & 2.75 \\
 & Mistral  & 1.76 & 1.97 & 1.79 & 2.75 & 3.02 & 3.60 \\
 & Qwen2.5  & 2.03 & 2.14 & 2.19 & 2.31 & 3.06 & 3.50 \\
 & Llama    & 1.68 & 1.97 & 1.85 & 2.83 & 3.57 & 3.72 \\
\midrule
\multirow{4}{*}{\rotatebox{90}{\scriptsize Onion}}
 & Qwen2-VL & 2.01 & 1.96 & 2.13 & 1.86 & 2.38 & 2.48 \\
 & Mistral  & 2.09 & 2.14 & 1.98 & 2.45 & 3.02 & 3.32 \\
 & Qwen2.5  & 2.28 & 2.28 & 2.32 & 2.19 & 3.09 & 3.27 \\
 & Llama    & 2.01 & 2.33 & 2.13 & 2.91 & 3.80 & 3.60 \\
\bottomrule
\end{tabular}
\caption{Mean judge scores (1--5) for the baseline (\textsc{base}), unreliable narrator (\textsc{un}) and structured control (\textsc{str}) prompts, without sentence classification. Malformed generations excluded as in Table~\ref{tab:main}.}
\label{tab:structured}
\end{table}

\section{LLM as Judge}
\label{appendix_llm_judge}
Each dimension is scored in a separate call to the LLM, so the judge does not see more than one criterion at a time. The judge is run at temperature 0 and receives only the headline, body text and the explanation to be evaluated. We separately score satire target and explanation (Table \ref{tab:satire_rubric}), since a model can provide a confident explanation built on a hallucinated target. UN conditions are additionally scored on trope\_accuracy (0-1), surface\_account, implied\_author and mechanism which are each (1-5). 

\section{Qualitative Analysis}
\label{appendix_qualitative}

% Our analysis shows that, for an Onion article with the headline ``Disappointed Man Reaches Bottom Of Ice Cream Carton Right When He Was Hitting His Stride", the model describes the fictional man and his ice cream carton as if they were a literal fact and misses the actual target being overeating. Almost half of the responses score 1 on satire target for this reason, pointing to future work in improving the retrieval scaffold. The scoring rubric is in Table~\ref{tab:satire_rubric} and a comparison of qualitative outputs in Table~\ref{tab:score_difference_comparison}

The scoring rubric is in Table~\ref{tab:satire_rubric}. Tables \ref{tab:score_difference_comparison}, \ref{tab:qual_books}, \ref{tab:qual_olympics} and \ref{tab:qual_icecream} show four qualitative examples. 

Table \ref{tab:score_difference_comparison} shows the pattern where under the baseline prompt the model names the target and stops, while the unreliable narrator prompt produces the trope, the surface account, the implied author's belief and the gap between them from the same knowledge.

Table \ref{tab:qual_books} shows a second comparison in which the unreliable narrator response raises the explanation score from 2 to 5, but its satire target score falls from 3 to 1 as the model constructs a biography for Bianca Sutton, including a follower count and an employment history, for a person the article invented.

Table \ref{tab:qual_olympics} shows the same issue in a single response. The model produces a good satire explanation and scores 4, but accepts the article's premise that Canada is a minnow nation, meaning an underdog candidate, and reports a real athlete and a real event as though the article were straightforwardly describing them leading to scoring 1 on satire target.

Table \ref{tab:qual_icecream} shows the limit of the framing as both prompts restate Billy Crenshaw and his ice cream carton as the satire target, missing the phenomenon the article is mocking leading to a score of 1. The unreliable narrator prompt still produces a structured explanation, but it applies that structure to the fiction which shows that the framing organises the explanation without improving the grounding that it relies on. Roughly half of all satire target scores are like this, which shows that the retrieval scaffold should be a target for future work.

\begin{table}[h]
\footnotesize
\centering
\begin{tabular}{@{} p{0.94\linewidth} @{}}
\toprule
\textbf{Input Headline:} ``Struggling Winter Olympics Minnow Nation Surprises With First Gold Medal'' \\
\midrule
\textbf{Satire target (1 / 5).} Mikaël Kingsbury, a Canadian freestyle skier, won the gold medal in men's moguls at the 2026 Winter Olympics in Milano Cortina, Italy. \\
\addlinespace[4pt]
\textbf{Explanation (4 / 5).} The Betoota Advocate employs the naif narrator trope to present a heartwarming story of a struggling nation, Canada, finally achieving a rare Winter Olympics victory. The implied author, however, recognises the true significance of Kingsbury's achievement, as he is one of the most decorated athletes in the history of moguls skiing. The gap between the narrator's portrayal of a fairy tale underdog story and the implied author's recognition of Kingsbury's true talent and legacy creates a satirical effect. \\
\bottomrule
\end{tabular}
\caption{An unreliable narrator output in which the explanation is coherent and scores 4, while the satire target scores 1. The model names a real athlete and a real event, but accepts the article's premise that Canada is a minnow nation rather than recognising it as the joke.}
\label{tab:qual_olympics}
\end{table}
\begin{table}[h]
\centering
\small
\setlength{\tabcolsep}{3.5pt}
\begin{tabular}{llcccc}
\toprule
 & & \multicolumn{2}{c}{\textit{target}} & \multicolumn{2}{c}{\textit{expl.}} \\
\cmidrule(lr){3-4}\cmidrule(lr){5-6}
Model & Prompt & raw & +\textsc{sc} & raw & +\textsc{sc} \\
\midrule
\multicolumn{6}{l}{\textit{Betoota Advocate} (AU)} \\
\multirow{2}{*}{Qwen2-VL-7B}
 & \textsc{base} & \textbf{2.00} & 1.64 & 2.00 & \textbf{2.04} \\
 & \textsc{un}   & \textbf{1.93} & 1.74 & 2.46 & \textbf{2.48} \\
\multirow{2}{*}{Mistral-7B}
 & \textsc{base} & \textbf{1.76} & 1.55 & 2.75 & \textbf{3.13} \\
 & \textsc{un}   & \textbf{1.97} & 1.61 & \textbf{3.02} & 2.98 \\
 \multirow{2}{*}{Qwen2.5-7B}
 & \textsc{base} & \textbf{2.03} & 1.93 & 2.31 & 2.31 \\
 & \textsc{un}   & \textbf{2.14} & 2.10 & 3.06 & \textbf{3.07} \\
\multirow{2}{*}{Llama-3.1-8B}
 & \textsc{base} & 1.68 & \textbf{1.69} & 2.83 & \textbf{3.04} \\
 & \textsc{un}   & \textbf{1.97} & 1.85 & \textbf{3.57} & 3.53 \\
\midrule
\multicolumn{6}{l}{\textit{The Onion} (US)} \\
\multirow{2}{*}{Qwen2-VL-7B}
 & \textsc{base} & \textbf{2.01} & 1.64 & \textbf{1.86} & 1.81 \\
 & \textsc{un}   & \textbf{1.96} & 1.79 & \textbf{2.38} & 2.34 \\
\multirow{2}{*}{Mistral-7B}
 & \textsc{base} & \textbf{2.09} & 1.54 & 2.45 & \textbf{2.68} \\
 & \textsc{un}   & \textbf{2.14} & 1.63 & 3.02 & 3.02 \\
\multirow{2}{*}{Qwen2.5-7B}
 & \textsc{base} & \textbf{2.28} & 1.96 & 2.19 & \textbf{2.30} \\
 & \textsc{un}   & \textbf{2.28} & 2.11 & 3.09 & \textbf{3.10} \\
\multirow{2}{*}{Llama-3.1-8B}
 & \textsc{base} & \textbf{2.01} & \textbf{2.01} & 2.91 & \textbf{2.99} \\
 & \textsc{un}   & \textbf{2.33} & 2.10 & \textbf{3.80} & 3.74 \\
\bottomrule
\end{tabular}
\caption{Effect of the sentence-classification pre-pass (+\textsc{sc}) against
raw article text. Bold marks the better of each pair.}
\label{tab:sc}
\end{table}

\begin{table*}[t]
\centering
\footnotesize
\renewcommand{\arraystretch}{1.2}
\begin{tabular}{cp{6.5cm}p{6.5cm}}
\toprule
\textbf{Score} & \textbf{Satire Target Score (\texttt{satire\_target})} & \textbf{Explanation Score (\texttt{satire\_explanation})} \\
\midrule
\textbf{1} & No real satire target identified, or entirely wrong / hallucinated. & Irrelevant, hallucinated, fails to explain, or merely restates the headline. \\
\textbf{2} & Correct general topic, but wrong specific people, dates, or details. & Identifies the target, but fails to explain how or why it is satirical. \\
\textbf{3} & Correct target with minor factual inaccuracies (e.g., wrong date, right person). & Explains the mechanism, but misses important cultural or political nuance. \\
\textbf{4} & Correct target with all key people and context; trivial differences only. & Explains target, mechanism, and most cultural context with minor gaps. \\
\textbf{5} & Fully accurate and specific: correct target (including, as applicable, event, people, dates, and AU context.) & Complete explanation: target, satirical device, and cultural critique. \\
\bottomrule
\end{tabular}
\caption{Evaluation rubric for satirical article comprehension across satire target grounding and satirical mechanism explanation.}
\label{tab:satire_rubric}
\end{table*}

\begin{table*}[t]
    \footnotesize
    \centering
    \begin{tabular}{@{} p{0.47\linewidth} p{0.47\linewidth} @{}}
    \toprule
    \multicolumn{2}{@{}l}{\textbf{Input Headline:} ``Corporations That Plaster Themselves In Rainbows Every Year Not Keen To Bail Out Mardi Gras Sadly''} \\
    \midrule
    \textbf{Baseline Prompt} & \textbf{Unreliable Narrator} \\
    \textbf{Score: 2 / 5} & \textbf{Score: 5 / 5} \\
    \addlinespace[4pt]
    The satire is satirizing the hypocrisy of large corporations, such as banks, tech companies, and food giants, who use the pride flag for branding purposes but are unwilling to support the Mardi Gras event financially. The article suggests that these corporations are more interested in profit than in genuine support for the LGBTQ+ community. 
    & 
    The Betoota Advocate employs the \textbf{naif narrator trope} to present the situation as a \textbf{simple failure of corporate sponsorship}, while the \textbf{implied author} is likely satirizing the hypocrisy of corporations using rainbow-themed branding for profit while not actually supporting the community. \textbf{The gap} between the narrator's surface account and the implied author's critique of performative rainbow capitalism creates the satirical effect. \\
    \addlinespace[6pt]
    \textit{Names the target, then stops without explaining the satirical device} & \textit{Names all four parts: trope, surface, author, gap} \\
    \bottomrule
    \end{tabular}
    \caption{Comparison of qualitative outputs and evaluation scores between the baseline prompt and the unreliable narrator framework.}
    \label{tab:score_difference_comparison}
\end{table*}

\begin{table*}[t]
\footnotesize
\centering
\begin{tabular}{@{} p{0.47\linewidth} p{0.47\linewidth} @{}}
\toprule
\multicolumn{2}{@{}l}{\textbf{Input Headline:} ``Girlfriend Takes A Pause From Scrolling To Declare She's Going To Read 20 Books This Year''} \\
\midrule
\textbf{Baseline Prompt} & \textbf{Unreliable Narrator} \\
\textbf{Target: 3 / 5 \quad Explanation: 2 / 5} & \textbf{Target: 1 / 5 \quad Explanation: 5 / 5} \\
\addlinespace[4pt]
\textit{Satire target.} The article is a satire about a woman named Bianka Sutton who claims she will read 20 books this year, but instead of actually reading, she continues to scroll through her TikTok feed.
&
\textit{Satire target.} Bianca Sutton is a social media influencer with over 38K followers on Instagram, where she posts content about her personal life, including her relationship with her boyfriend, Peter Evans. She is also a professional in the recruitment industry, as indicated by her LinkedIn profile. \\
\addlinespace[4pt]
\textit{Explanation.} The satire is about the superficial nature of New Year's resolutions and the tendency to prioritize social media over personal growth.
&
\textit{Explanation.} The Betoota Advocate employs the \textbf{naif narrator trope} in this article. The \textbf{surface account} presents Bianca Sutton as a woman who has decided to break away from her TikTok addiction and read 20 books this year. However, the \textbf{implied author}, who is likely a satirist, actually sees this as a continuation of her usual behaviour, where she spends most of her time on social media. \textbf{The gap} between the narrator's presentation of a serious change in behaviour and the implied author's recognition of it as a superficial and ultimately unchanged pattern creates the satirical effect. \\
\addlinespace[6pt]
\textit{Names the target but does not explain the device; summarises the article's plot rather than the phenomenon.} & \textit{Names all four parts, but invents a real-world identity for a fictional character.} \\
\bottomrule
\end{tabular}
\caption{A second qualitative comparison. The unreliable narrator prompt raises explanation from 2 to 5, but satire target falls from 3 to 1 because the model fabricates biographical detail for a character the article invented.}
\label{tab:qual_books}
\end{table*}

\begin{table*}[t]
\footnotesize
\centering
\begin{tabular}{@{} p{0.47\linewidth} p{0.47\linewidth} @{}}
\toprule
\multicolumn{2}{@{}l}{\textbf{Input Headline:} ``Disappointed Man Reaches Bottom Of Ice Cream Carton Right When He Was Hitting His Stride'' (The Onion)} \\
\midrule
\textbf{Baseline Prompt} & \textbf{Unreliable Narrator} \\
\addlinespace[4pt]
\textit{Satire target.} Billy Crenshaw, a local man in Clemson, SC, is disappointed after reaching the bottom of an ice cream carton just as he was hitting his stride.
&
\textit{Satire target.} A man, Billy Crenshaw, is disappointed after finishing an ice cream carton because he had just started feeling satisfied with the dessert. \\
\addlinespace[4pt]
\textit{Explanation.} The article is satirizing the idea that people can become so engrossed in eating ice cream that they become disappointed when they finish the carton.
&
\textit{Explanation.} The satire uses the \textbf{naif narrator trope}, presenting Billy Crenshaw's disappointment as a serious and relatable experience. The \textbf{implied author}, however, is likely to see the situation as humorous and exaggerated. \textbf{The gap} between the narrator's surface account of Billy's disappointment and the implied author's recognition of the absurdity of his reaction creates the satirical effect. \\
\addlinespace[6pt]
\multicolumn{2}{@{}p{0.96\linewidth}}{\textit{Judge feedback on the satire target:} ``Restates the fictional character and his ice cream carton as if it were literal fact, instead of identifying the real satirical target: sports-metaphor language applied to overeating.''} \\
\bottomrule
\end{tabular}
\caption{Both prompts restate the article's invented character as the satire target and score 1 on that dimension. The unreliable narrator prompt still produces a structured explanation, but the structure is applied to the fiction rather than to the phenomenon the article is mocking.}
\label{tab:qual_icecream}
\end{table*}

\end{document}